\documentclass[letterpaper, 10 pt, conference]{ieeeconf}
\IEEEoverridecommandlockouts
\usepackage{cite}
\usepackage{amsmath,amssymb,amsfonts}
\usepackage{algorithmic}
\usepackage{graphicx}
\usepackage{textcomp}
\usepackage{xcolor}
\usepackage{float}
\usepackage{makecell}
\usepackage{hyperref}
\usepackage{caption}
\def\BibTeX{{\rm B\kern-.05em{\sc i\kern-.025em b}\kern-.08em
    T\kern-.1667em\lower.7ex\hbox{E}\kern-.125emX}}
\begin{document}

\title{\LARGE \bf MULE – Multi-terrain and Unknown Load Adaptation  for Effective Quadrupedal Locomotion}

\author{
Vamshi Kumar Kurva$^{1}$, Shishir Kolathaya$^{2}$
\thanks{$^{1}$VK. Kurva is with the Department of Computer Science \& Automation, Indian Institute of Science, Bengaluru.}%
\thanks{$^{2}$S. Kolathaya is with the Centre for Cyber-Physical Systems and the Department of Computer Science \& Automation, Indian Institute of Science, Bengaluru.}
\thanks{This work is supported by ARTPARK and ADB.}
\thanks{Email: \href{mailto:stochlab@iisc.ac.in}{stochlab@iisc.ac.in}, Project Website: \href{https://www.stochlab.com/MULE/}{stochlab.com/MULE}}
}

\maketitle

\begin{abstract}
Quadrupedal robots are increasingly deployed for load-carrying tasks across diverse terrains. While Model Predictive Control (MPC)-based methods can account for payload variations, they often depend on predefined gait schedules or trajectory generators, limiting their adaptability in unstructured environments. To address these limitations, we propose an Adaptive Reinforcement Learning (RL) framework that enables quadrupedal robots to dynamically adapt to both varying payloads and diverse terrains. The framework consists of a \textit{nominal policy} responsible for baseline locomotion and an \textit{adaptive policy} that learns corrective actions to preserve stability and improve command tracking under payload variations. We validate the proposed approach through large-scale simulation experiments in Isaac Gym and real-world hardware deployment on a Unitree Go1 quadruped. The controller was tested on flat ground, slopes, and stairs under both static and dynamic payload changes. Across all settings, our adaptive controller consistently outperformed the controller in tracking body height and velocity commands, demonstrating enhanced robustness and adaptability without requiring explicit gait design or manual tuning.
\end{abstract}



\section{Introduction}
The load-carrying capability of quadrupedal robots is essential for enhancing their deployment across various domains, including logistics, search and rescue, military operations, and agriculture. Enabling these robots to transport substantial payloads can significantly improve operational efficiency, reducing the need for human intervention in hazardous or hard-to-reach environments. Although substantial advances have been made in legged locomotion, particularly in traversing uneven terrain and handling external disturbances, the challenge of adapting to unknown payloads remains relatively less explored.  

\begin{figure}[h]
    \centering
    \includegraphics[width=0.46\textwidth]{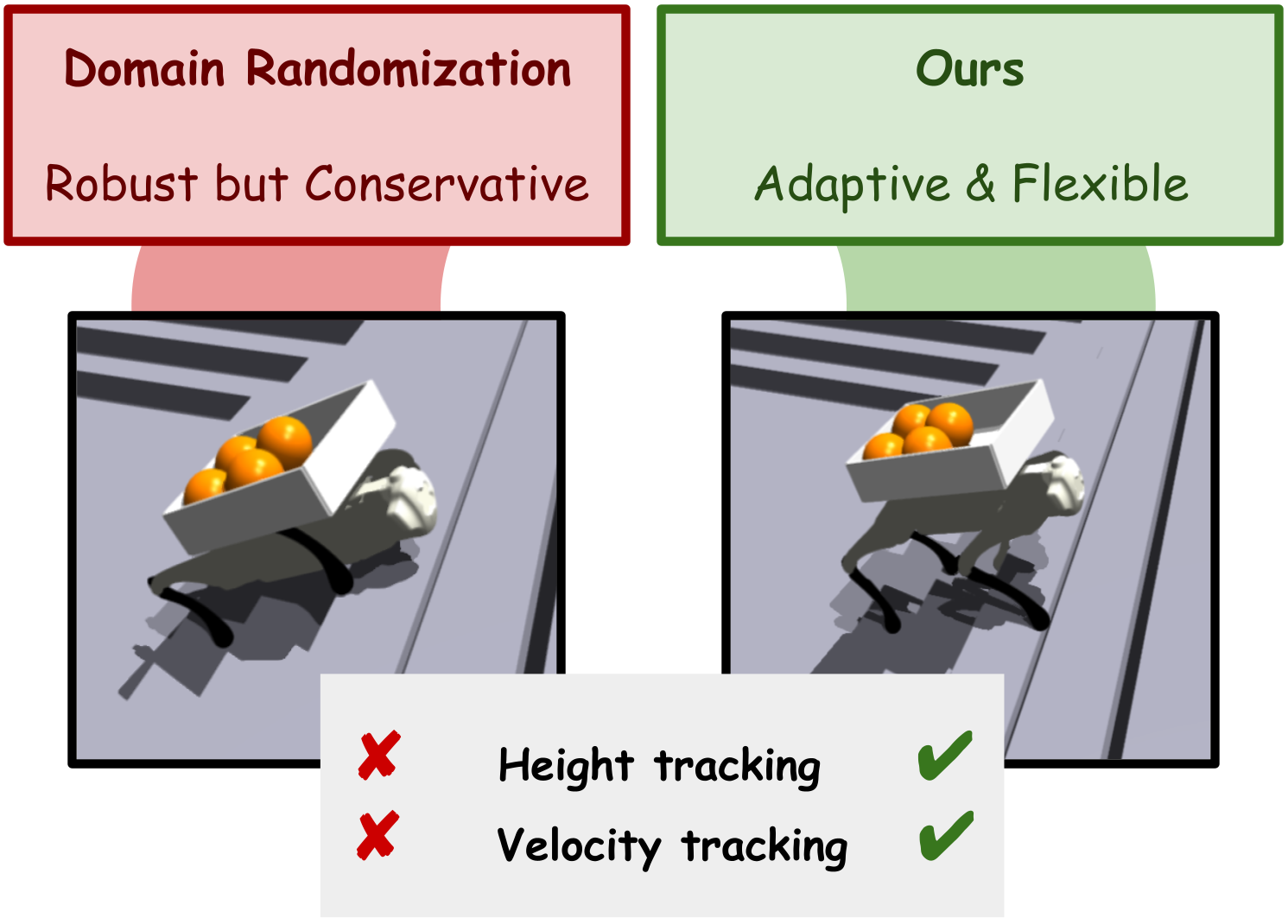}
    \caption{Domain Randomization vs. Adaptive RL: While domain randomization is robust but conservative, our adaptive approach enables flexible locomotion with improved height and velocity tracking across challenging terrains.
    }
    \label{fig:adapr_rl_small}
\end{figure}

Several studies have attempted to address this issue. \cite{Online_payload_iros_17} employs an online recursive method that estimates the robot’s inertial parameters and base center of mass (CoM) using contact forces and joint angles. However, this approach requires the robot to halt during payload detection, restricting its suitability for real-time applications. \cite{learning_unknown_dynamics} circumvents direct parameter identification by learning a locally linear, time-varying residual model around the current trajectory. This technique supports real-time control and demonstrates effective payload handling with a 10 kg load on a 12 kg A1 robot, though the assumption of symmetric load distribution minimizes CoM shifts.  

Other approaches focus on adaptive and robust control strategies. \cite{adaptive_force_control} integrates $\mathcal{L}_1$ adaptive control into a force control framework, enabling a Unitree A1 quadruped to stably transport a 6 kg payload. Similarly, \cite{RCMPC} introduces a robust min-max MPC strategy based on robust optimization to account for system uncertainties. \cite{Adaptive_clf_mpc} incorporates Control Lyapunov Function (CLF) constraints within an MPC framework to ensure stable and adaptive locomotion, with validation conducted on the ANYmal robot. Meanwhile, \cite{rl_augmented_mpc} integrates Reinforcement Learning with Model Predictive Control to achieve adaptive balancing and swing foot reflection, allowing quadrupedal robots to dynamically adjust to payload variations and external disturbances. Their framework successfully demonstrated payload handling of 7 kg on a Unitree Go1 robot on flat ground.

All the above-mentioned methods are model-based force controllers that regulate ground reaction forces (GRFs) at the stance feet based on desired height and payload variations. Also, all the above methods showed payload adaptation mostly on the flat terrain or smooth sloped terrains. These approaches typically model quadrupeds as a single rigid body (SRB) and compute the optimal GRFs to be applied at contact points, while a low-level PD controller tracks the swing leg trajectories. To enforce structured locomotion, they rely on gait or trajectory generators to predefine foot contact schedules based on gait and velocity, enforcing distinct control strategies for swing and stance legs. A switch-based controller applies force control to stance legs and PD control to swing legs, making the system sensitive to early or delayed contacts on unstructured terrains, which can induce instability. In contrast, RL-based approaches have demonstrated effective locomotion across unstructured terrains without relying on predefined gait schedules \cite{RMA} \cite{walk_these_ways} \cite{dreamwaq} \cite{eth_legged_gym} \cite{HimLoco} \cite{piploco}.  These methods directly output desired joint positions, which are tracked using a PD controller, eliminating the need for phase-based switching. By learning policies that implicitly adapt to terrain variations and contact conditions, RL-based controllers achieve more robust and versatile locomotion compared to model-based methods.

Building upon the strengths of RL-based methods for unstructured terrain locomotion, we propose an Adaptive RL framework that enables load-carrying capability across a variety of terrains. Unlike traditional model-based methods that rely on explicit model adjustments for varying loads, our approach allows the quadruped to dynamically adjust its locomotion strategy based on perceived changes in payload. This eliminates the reliance on predefined gait schedules, offering greater robustness and versatility in handling both terrain and payload variations. Our key contributions are summarized as follows:
\begin{itemize} \item We introduce an \textbf{Adaptive RL framework} for locomotion under varying payload conditions by augmenting a nominal policy with an adaptive corrective policy.
\item The framework is trained in a two-phase process: the nominal policy is first trained under normal conditions, followed by an adaptive policy that provides corrective actions without the need for explicit payload parameter estimation.
\item We demonstrate that the adaptive policy significantly improves command tracking in payload scenarios, especially on challenging terrains such as stairs with added payloads.
\item The proposed framework is validated in simulation and hardware, showing notable improvements compared to the baseline approach. \end{itemize}

\section{Methodology}

\subsection{Preliminaries}
Reinforcement Learning (RL) provides a framework for training autonomous agents to maximize cumulative rewards in an environment by interacting with it through trial and error. In the context of quadruped locomotion, RL formulates the control problem as a Markov Decision Process (MDP), where the robot learns an optimal policy to achieve stable and efficient movement across diverse terrains. The MDP is defined by a tuple \((\mathcal{S}, \mathcal{A}, \mathcal{P}, r, \gamma)\), where: \(\mathcal{S}\) represents the state space, \(\mathcal{A}\) defines the action space, \(\mathcal{P}\) denotes the transition dynamics, which model how the quadruped's state evolves based on applied actions and environmental interactions, \(r: \mathcal{S} \times \mathcal{A} \to \mathbb{R}\) is the reward function and \(\gamma \in (0,1]\) is the discount factor that balances immediate and long-term rewards.

The goal of reinforcement learning is to learn a policy \(\pi_{\theta}(a_t \mid s_t)\), parameterized by \(\theta\), which defines the probability of selecting action \(a_t \in \mathcal{A}\) given the current state \(s_t \in \mathcal{S}\). The agent interacts with the environment over discrete time steps \(t = 0, 1, 2, \dots\), receiving a reward \(r(s_t, a_t)\) at each step based on the current state and action. The objective of RL is to learn a policy that maximizes the expected cumulative discounted reward:
\[
J(\pi) = \lim_{T \to \infty} \mathbb{E}\left[\sum_{t=0}^T \gamma^t r(s_t, a_t)\right]
\]

\begin{figure*}[ht]
	\centering	\includegraphics[width=0.9\linewidth]{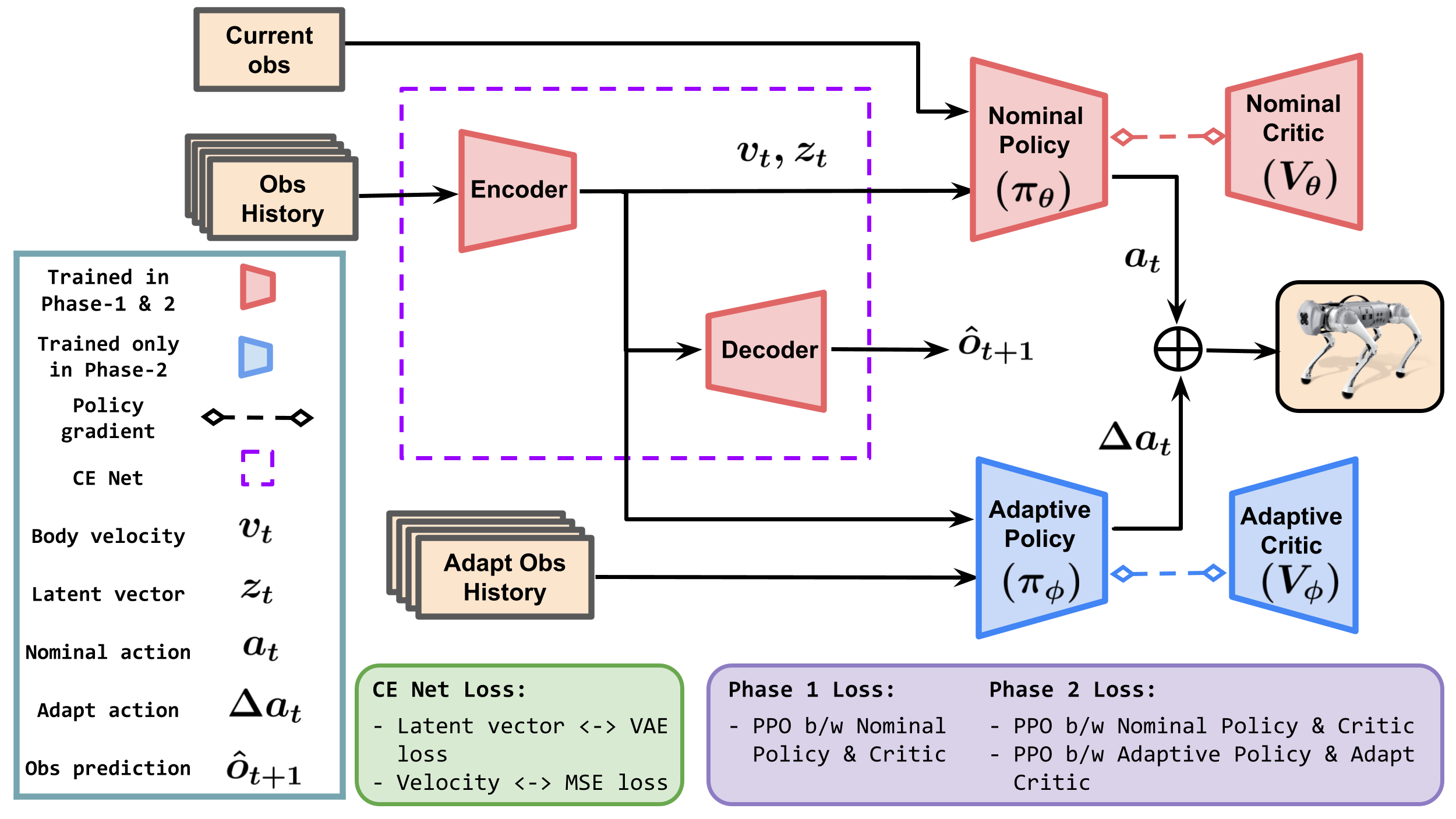}
	\caption{\textbf{Overview of the proposed framework} - History of observations is encoded to get a latent vector and body velocity using CE Net. The nominal policy and critic are trained in Phase 1, while the adaptive policy and adaptive critic are introduced in Phase 2 to enhance adaptation to payload variations. The combined action enables robust locomotion across diverse payload conditions.} 
	\label{fig:adaptive_rl}		
\end{figure*}

\subsection{Background}

The evolution of RL in quadrupedal locomotion has progressed significantly, marked by innovative methodologies and frameworks. A foundational work from ETH Zurich was the first to utilize Nvidia's Isaac Gym, a high-performance GPU-accelerated simulator, to train quadrupedal robots like ANYmal \cite{eth_legged_gym}. This approach employed a teacher-student architecture in which the teacher policy, trained with privileged information, could not be directly deployed on hardware due to its reliance on this information. Instead, a student policy was developed to infer latent variables corresponding to the privileged information from the observation history.  

Following this initial framework, the RMA algorithm emerged, focusing on real-time adaptation in quadrupedal robots without relying on domain knowledge or reference trajectories \cite{RMA}. Building on this foundation, Walk These Ways \cite{walk_these_ways} explored achieving a multiplicity of behaviors through a single policy by manipulating gait-based rewards, enhancing locomotion versatility.  

The introduction of DreamWaQ \cite{dreamwaq} marked a significant milestone as it utilized an Asymmetric Actor Critic architecture, demonstrating the ability to develop robust and generalized policies capable of navigating diverse terrains such as rough slopes and stairs. This work highlighted the potential of RL to create adaptable locomotion strategies that can handle complex environments effectively.  

Domain randomization techniques are commonly used to bridge the gap between simulation and real-world deployment by introducing small variations in robot parameters during training. This helps develop policies that are resilient to disturbances. However, when the range of parameter variations is too large, the resulting policies tend to be overly conservative \cite{DR_with_entropy_maximisation}\cite{Revisiting_DR_RA_PPO}, prioritizing robustness at the cost of optimal performance. These limitations highlight the need for adaptive policies that can dynamically adjust to varying conditions, such as payload changes, rather than relying on a one-size-fits-all approach.

\subsection{Adaptive RL Framework}

When a payload is added or removed from the robot, it causes significant changes in system parameters such as mass, center of mass (CoM), and inertia, altering the dynamics of the system. Explicitly estimating these parameters in real time can be challenging and error prone. Instead, we propose an adaptive framework where a corrective action is learned to compensate for these changes.  

Our proposed \textbf{Adaptive RL framework} (Fig. \ref{fig:adaptive_rl}) is inspired by adaptive control methods from classical control and recent RL works \cite{robustmodelbasedrl} \cite{safereinforcementlearningdual}. It consists of two phases of training:  
\begin{itemize}
    \item \textbf{Phase 1:} We train a nominal policy under normal conditions (without payload). The nominal policy is responsible for basic locomotion and command tracking in standard scenarios.  
    \item \textbf{Phase 2:} We train an adaptive policy to provide corrective actions under payload conditions, treating the added payload as an external disturbance. The adaptive policy works alongside the nominal policy to ensure that the robot maintains desired command tracking, such as base height and velocity, even under varying payloads.  
\end{itemize}

Instead of explicitly estimating the unknown payload parameters, the adaptive policy dynamically adjusts the actions to compensate for their effects. This approach allows the quadruped to adapt its locomotion strategy without predefined gait schedules or explicit model adjustments, offering greater robustness and versatility in handling terrain and load variations.

\subsection{Phase-1: Nominal Policy Training}
The objective of Phase-1 is to train the nominal policy for robust locomotion across diverse terrains.

\textbf{Observations} - 
The observation is a $45$ dimensional vector consisting of the following
\begin{equation*}
    o_t = \begin{bmatrix}
        \omega_t & g_t & c_t & \theta_t & \dot{\theta}_t & a_{t-1}
    \end{bmatrix}^T
\end{equation*}
where $w_t, g_t$ are the body angular and gravity vectors, $c_t$ is the body velocity commands, $\theta_t \dot{\theta_t}$ are the joint angle positions and velocities, $a_{t-1}$ is the previous action. 

\textbf{Actions} -
The action is $12$ dimensional vector which represents the desired joint positions relative to a fixed standing pose $\theta_{stand}$, i.e. 
\[ \theta_{des} = \theta_{stand} + a_t\] 
The desired joint angles are tracked using a PD controller. 

\textbf{Rewards} -
Total reward at any given time-step is given by
\[ r_t(\theta) = \sum_i r_i w_i\]
where $\theta$ is the parametrization of the nominal policy, $i$ is the index of the reward component and $w_i$ is the weight as shown in Table \ref{tab:reward_table}

\textbf{Encoder} -
The encoder is a part of Context Estimator Network (CE Net) that encodes the history of observations $o_t^{H}$ into a latent vector $z_t$ and body velocity $v_t$. A decoder is used to reconstruct the next observations from the encoding. $\beta$-VAE is used for this auto-encoding task.  CE Net is optimized using a hybrid loss
function, defined as follows:
\[ \mathcal{L}_{CE} = \mathcal{L}_{est} + \mathcal{L}_{VAE}\]
These losses are taken directly from \cite{dreamwaq}.

\textbf{Training} -
We denote the \textbf{Nominal Policy} as $\pi_\theta$ with observation $o_t$ and action $a_t$, while the \textbf{Adaptive Policy} $\pi_\phi$ takes an augmented observation $\tilde{o}_t$ and outputs the corrective action $\Delta a_t$. In the first phase, we train only the nominal policy $\pi_\theta$ using Proximal Policy Optimization (PPO) \cite{PPO}, while the adaptive policy $\pi_\phi$ remains inactive (i.e., $\nabla_{\phi} = 0$). The final action applied to the environment is $a_t$. The PPO objective for the nominal policy is:
\begin{equation*}
    \mathcal{L}_{\text{PPO}}^{\theta} = \mathbb{E} \left[ \min \left( \rho_t(\theta) \hat{A}_t^\theta, \ \text{clip}\left(\rho_t(\theta), 1-\epsilon, 1+\epsilon\right) \hat{A}_t^\theta \right) \right],
\end{equation*}
where $\rho_t(\theta)$ is the probability ratio between the current and old policies:
\begin{equation*}
    \rho_t(\theta) = \frac{\pi_\theta(a_t \mid o_t)}{\pi_{\theta_{\text{old}}}(a_t \mid o_t)},
\end{equation*}
and $\hat{A}_t^\theta$ is the advantage estimate for the nominal policy.

    

\subsection{Phase-2: Adaptive Policy Training under Varying Payloads}
In Phase-2, we train both the nominal policy and the adaptive policy simultaneously to handle varying payload conditions. While the nominal policy retains the same reward structure from Phase-1, the adaptive policy is optimized with a separate reward formulation that prioritizes stability, base height regulation, and load adaptation.

The primary objective of the adaptive policy is to maintain the robot’s desired base height under varying payloads. When the base height drops below the target due to an increase in payload, the policy needs to apply greater forces at the stance feet to restore it. Estimating end-effector forces (foot forces) is crucial for achieving this corrective behavior. We estimate these forces at each foot using the Jacobian relationship between the applied joint torques and the resulting forces:
\begin{equation*}
    \tau = J(\theta)^T f \implies f = \left(J(\theta)^T\right)^{\dagger} \tau, 
\end{equation*}
where \( J \) is the Jacobian matrix that depends on the joint configuration \(\theta\), \(\tau\) is the vector of applied joint torques, and \( f \) represents the estimated end-effector forces.
To incorporate this information into the adaptive policy, we augment its observation space with the estimated foot forces:
\begin{equation*}
    \text{Adapt obs } \tilde{o}_t = (\text{obs}, f).
\end{equation*}
We introduce a GRF tracking reward to encourage the adaptive policy to generate higher GRFs when the base height falls below the desired target.
\begin{equation*}
\begin{aligned}
    r_{\text{GRF}} =\ & 0.75 \times (h > h_{cmd}) \\
    & + 0.50 \times (h < h_{cmd}) \times 
    \left( \sum_{i=1}^4 \lvert f_i \rvert > (m_r + m_p)g \right)
\end{aligned}
\end{equation*}

where \( h \) is the current base height, \( h_{cmd} \) is the desired base height, \( f_i \) is the GRF at leg \( i \), and \( m_r \) and \( m_p \) are the masses of the robot and payload, respectively.

We introduce dynamic payload variations to train the robot to adapt to changing mass conditions. A lightweight tray (250 g) is mounted on the robot’s base, and at the start of each episode, four spherical objects (balls) are placed inside it. The initial mass of each ball is sampled from a uniform distribution \([0, 1]\) kg, resulting in a total payload of up to 4 kg. The mass of each ball is resampled every 4 seconds from a uniform distribution \([0, 2.5]\) kg. This continuous variation forces the robot to adapt its actions in response to changing dynamics, ensuring stability and accurate command tracking. 

In this phase, the final action applied to the environment is $a_t + \Delta a_t$. The PPO objective for each policy is given by:
\begin{equation*}
    \mathcal{L}_{\text{PPO}}^\psi = \mathbb{E} \left[ \min \left( \rho_t(\psi) \hat{A}_t^{\psi}, \ \text{clip}\left(\rho_t(\psi), 1-\epsilon, 1+\epsilon\right) \hat{A}_t^{\psi} \right) \right],
\end{equation*}
where $\rho_t(\psi)$ is the probability ratio between the current and old policies ($\psi \in \{\theta, \phi\}$), and $\hat{A}_t^{\psi}$ is the corresponding advantage estimate. 

The gradient updates for Phase 2 are given by:  
\begin{align*}
    \theta &\gets \theta - \eta \nabla_{\theta} \mathcal{L}_{\text{PPO}}^\theta, \\ 
    \phi &\gets \phi - \eta \nabla_{\phi} \mathcal{L}_{\text{PPO}}^\phi.
\end{align*} 

\begin{table}[h]
    \vspace{-2mm}
    \centering
    \small
    \begin{tabular}{l c c}
        \hline
        \textbf{Reward} & \textbf{\makecell{Nominal \\ weights $(w_i)$}} & \textbf{\makecell{Adaptive \\ weights $(\alpha_i)$}} \\
        \hline
        Linear velocity tracking & \( 1.0\) & 0.0 \\
        Angular velocity tracking & \( 0.5 \) & 0.0 \\
        Linear velocity (\(z\)) & \( -2.0 \) & -2.0 \\
        Angular velocity (\(xy\)) & \( -0.05 \) & -0.05 \\
        Orientation & \( -0.2 \) & -0.2 \\
        Joint accelerations & \( -2.5 \times 10^{-7} \) & \(-2.5 \times 10^{-7}\) \\
        Joint power & \( -2.0 \times 10^{-5} \) & \(0.0\) \\
        Body height & \( -2.0 \) & -2.0 \\
        Foot clearance & \( -0.01 \) & -0.01 \\
        Action rate & \( -0.001 \) & -0.01 \\
        Smoothness & \( -0.01 \) & -0.01 \\
        GRF tracking & \( 0.0 \) & 2.0 \\
        \hline
    \end{tabular}
    \caption{Reward Weights for Nominal and Adaptive Policies: The nominal policy prioritizes velocity tracking, while the adaptive policy focuses on GRF tracking and body height stabilization}
    \vspace{-1em}
    \label{tab:reward_table}
\end{table}

\section{Results}
\subsection{Simulation}
We used the Isaac Gym simulator to validate our controller. The policy was trained using 4096 agents with a history size of $H = 5$ on a Nvidia RTX A6000 GPU. All actor and critic networks consist of three hidden layers with 512, 256, and 128 units, respectively. The encoder network has two hidden layers with 128 and 64 units, while the decoder network also contains two hidden layers with 64 and 128 units, respectively. During Phase-1, only the nominal policy was trained for 1000 iterations. In Phase-2, the nominal policy weights from Phase-1 were restored, and both policies were trained simultaneously for 500 iterations. Joint angles in simulation were tracked using an actuator network pre-trained for the Unitree Go1, ensuring realistic actuator behavior \cite{walk_these_ways}\cite{actuator_net}. We compared the command tracking performance against DreamWaQ, with the base mass randomized in the range 
$[0,10]$ kg, which we refer to as the \textbf{Baseline} controller.

\begin{figure*}[htp!]
    \centering
    \small
    \captionsetup{font=small}

    \includegraphics[height=0.4\linewidth, width=0.8\linewidth]{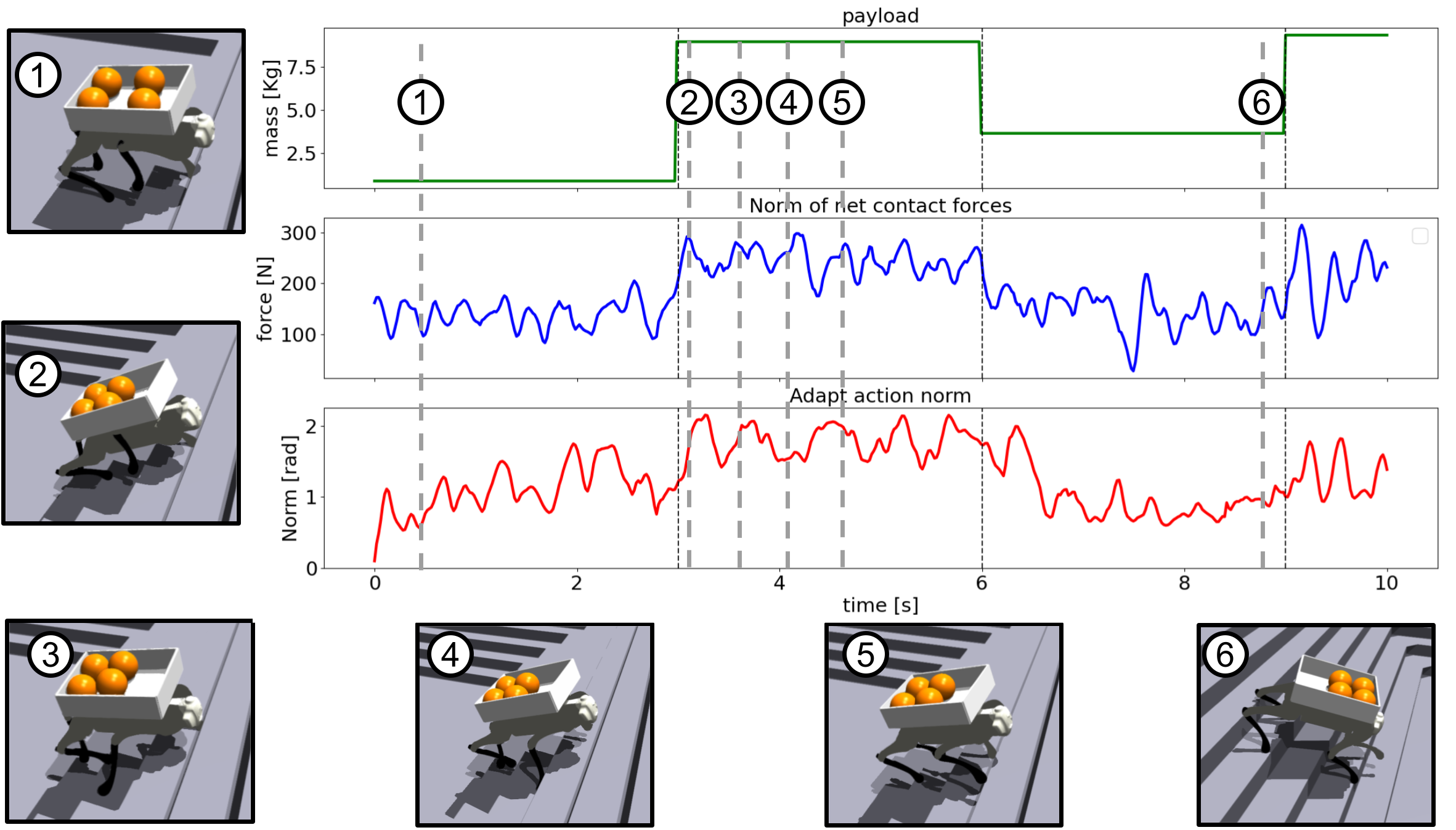}
    \caption{Adaptation of the quadruped robot to varying payload on stairs. (Top) The payload mass profile with 6 phases indicating mass transitions. (Middle) Norm of net contact forces over time. (Bottom) Norm of adaptive actions, demonstrating the controller’s response to mass changes and terrain transitions. Snapshots (1–6) depict representative instances during the locomotion sequence. (2), (3), (4) and (5) show how the robot has recovered from heavy payload change by generating higher GRFs. The graphs also show a positive correlation between the norm of the adapt action and the net contact forces.}
        \label{fig:key_frame}
\end{figure*}

\begin{figure}[htp!]
    \centering
    \small
    \captionsetup{font=small}    
    \includegraphics[width=0.98\columnwidth]{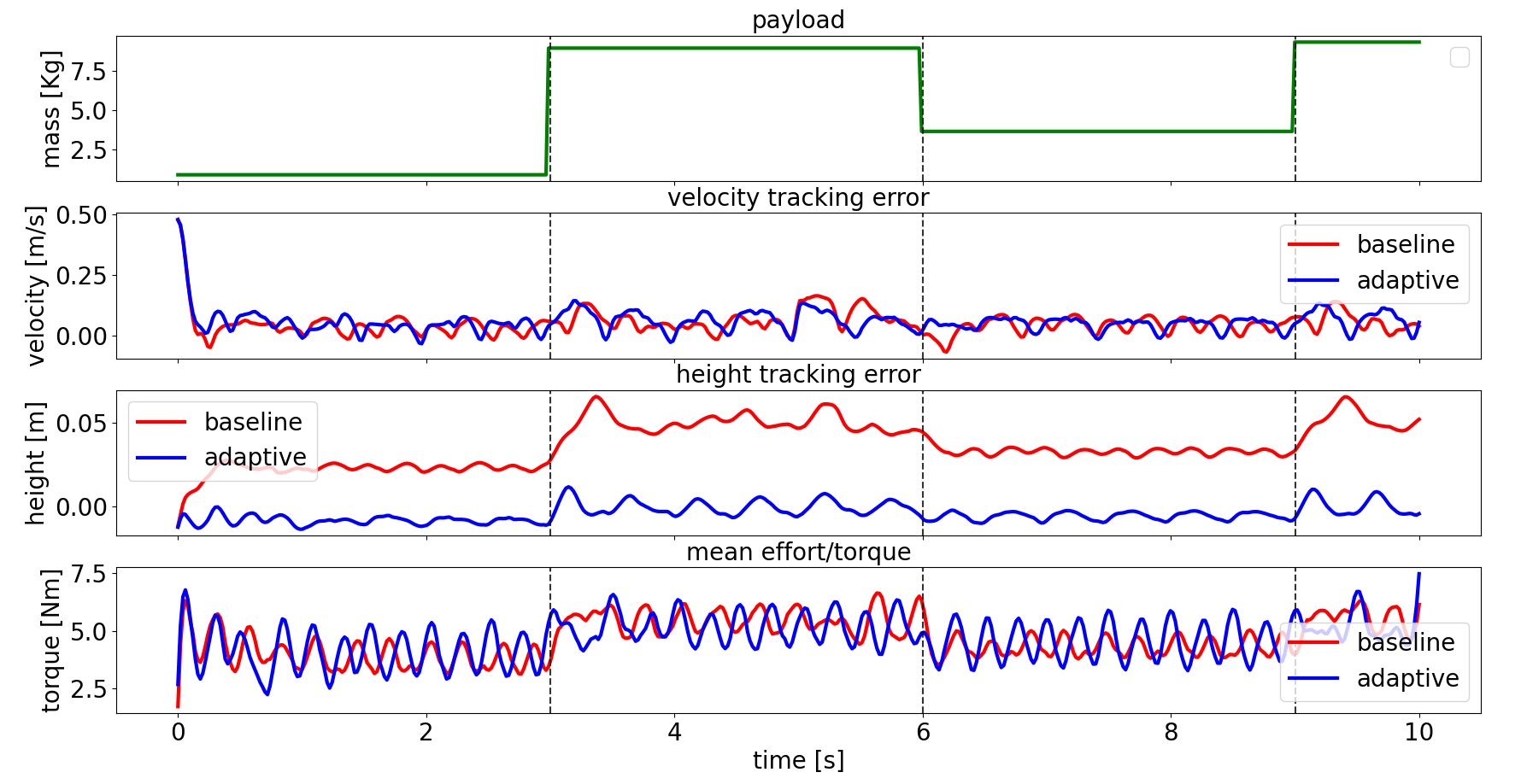}
    \caption{\textbf{Performance comparison of baseline and adaptive controllers on flat terrain}. The top section shows the payload mass profile. Both controllers exhibit similar velocity tracking and mean torque effort, while the adaptive controller achieves significantly better height tracking.}
    \label{fig:image1}
\end{figure}

\begin{figure}[htp!]
    \centering
    \small
    \captionsetup{font=small}
    \includegraphics[width=0.98\columnwidth]{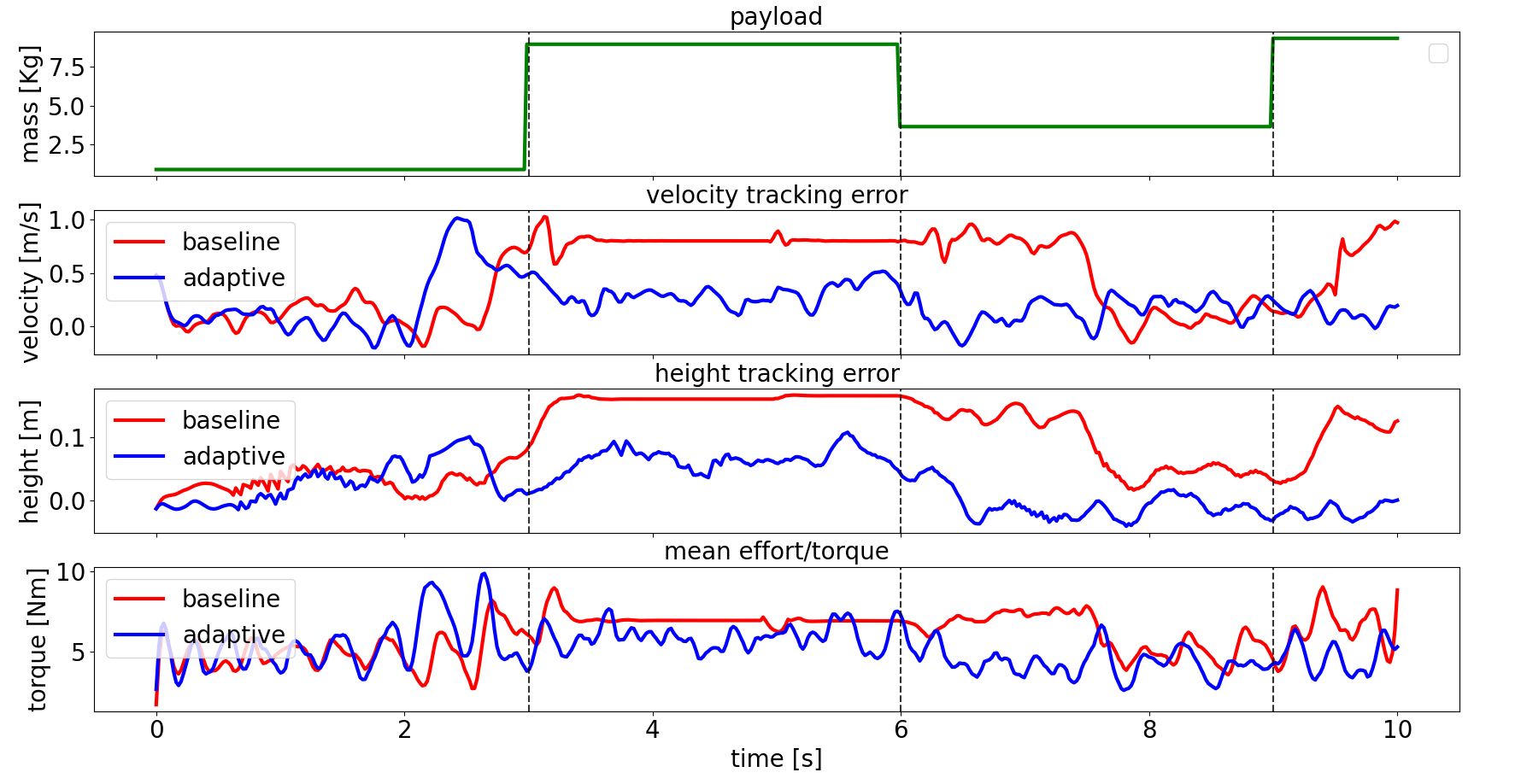}
    \caption{\textbf{Performance comparison of baseline vs adaptive controllers on stairs}. The payload mass profile is shown at the top. The flat red segments in the curves indicate instances where the baseline controller fails to adjust to sudden payload changes, causing the robot to come to a complete halt. In contrast, the adaptive controller maintains stable locomotion, achieving better velocity and height tracking while significantly reducing height tracking error and torque effort across different payload conditions.}
    \label{fig:image2}
\end{figure}

\begin{figure*}[htp!]
    \centering
    \small
    \captionsetup{font=small}
    \includegraphics[height=0.5\linewidth, width=0.9\linewidth]{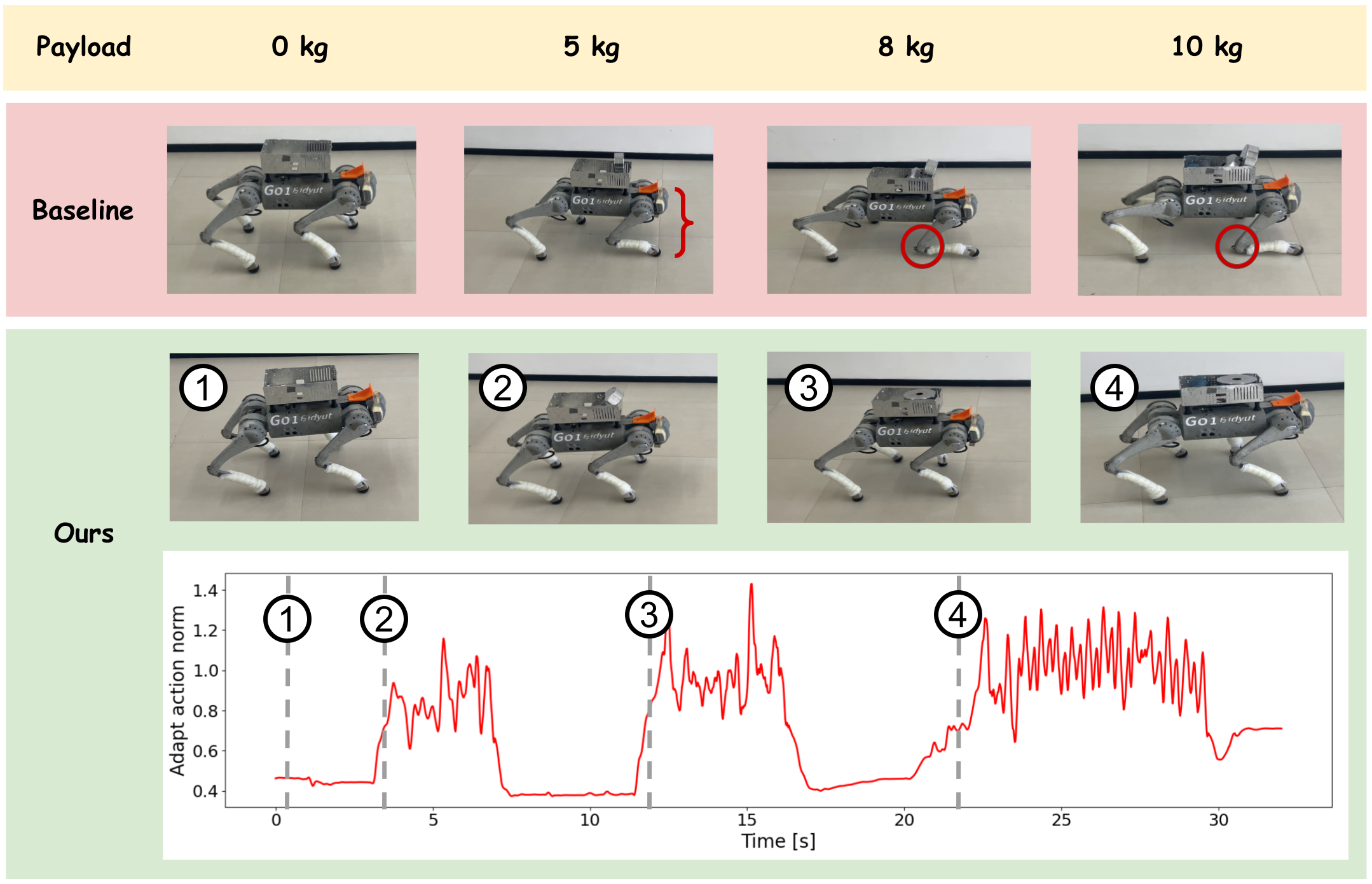}
    \caption{Comparison of quadruped locomotion performance under progressively increasing payloads (0-10 kg) using the baseline controller (top row) and the proposed adaptive controller (middle row). The bottom plot shows the adapt action norm over time, capturing the adaptive controller's response to payload changes. 
    Each numbered marker specifically denotes the moment additional payload was added to bring the total payload to the indicated value.
    Between two consecutive markers, the payload remains constant at the indicated value. The flat segments where the adapt action norm is close to zero correspond to brief halts where the robot was stopped to place or adjust the payload.}
    \label{fig:hw_key_frame}
\end{figure*}

\textbf{Flat Terrain} -
Figure \ref{fig:image1} shows the command tracking performance of baseline and adaptive policies on flat terrain under varying payloads. Although both policies track forward velocity commands with minimal error, the adaptive policy achieves significantly better height tracking accuracy. The baseline policy exhibits a strong positive correlation between height tracking error and payload, indicating its limited ability to generate sufficient GRFs as the payload increases.

\textbf{Stairs} -
Figure \ref{fig:image2} compares the performance of both policies on stairs. The baseline policy struggles to maintain velocity tracking under high payload conditions and eventually comes to a stop. Height tracking errors further indicate its inability to generate adequate GRFs to maintain the desired base height. In contrast, the adaptive policy tracks commands more effectively, outperforming the baseline policy.

Figure \ref{fig:key_frame} illustrates how adaptive actions and GRFs contribute to command tracking. As the payload increases, the adaptive policy increases both GRF magnitude and adaptive action norm, allowing the robot to maintain the desired height and follow velocity commands. This positive correlation demonstrates that the adaptive policy adjusts GRFs to compensate for higher payloads. Interestingly, although the adaptive policy is not explicitly instructed when to intervene, it learns through training to provide corrective actions precisely when the nominal policy struggles to track commands. This emergent behavior highlights the adaptive policy’s ability to address varying conditions based on its distinct objectives.

\subsection{Hardware deployment}
Real-world experiments were conducted on a Unitree Go1 robot, with a 500 g stainless steel tray mounted on its base. To simulate dynamic payload variations, multiple 1 kg iron balls were placed inside the tray, allowing them to move freely and induce shifts in the CoM. This introduced unpredictable payload dynamics throughout the experiment. Additionally, controlled static payload variations were tested by adding and removing 3 kg and 5 kg disks at different times during the trials. These variations allowed us to evaluate the policy’s ability to adapt to both gradual and sudden changes in payload conditions. Joint angles commanded by the policy are tracked using a PD controller with $k_p = 20.0 \text{ and } k_d=0.5$. Fig \ref{fig:hw_key_frame} shows the baseline vs adaptive controller under progressive increase of payload up to 10 kg. The baseline controller struggles to maintain stable locomotion, with noticeable foot scuffing and instability at higher payloads, while the adaptive controller successfully compensates for the added weight, maintaining balance and coordination. The bottom plot depicts the norm of the adaptive action output over time, illustrating the controller's response to payload changes. Each spike in the adaptation action norm corresponds to the moment a new payload is added, demonstrating the controller's ability to detect and respond to changes in mass and dynamics. We also evaluated the controller on slopes, and stairs and showed that the adaptive controller outperforms the baseline in all the cases.

\section{Conclusion}
We presented an RL-based adaptive control framework for quadrupedal locomotion across varying terrains and payload configurations. The proposed approach introduces an adaptive policy that provides corrective actions to complement the nominal policy, improving overall performance. Both policies are trained to optimize their respective rewards, with some common objectives, such as stability, shared between them. This shared reward structure fosters implicit cooperation, where the adaptive policy complements rather than conflicts with the nominal policy, injecting only the minimal corrective actions necessary to adapt to unexpected disturbances without overriding the baseline behavior. This modular design allows the system to retain the nominal policy’s learned behaviors under normal conditions while leveraging the adaptive policy’s rapid response capability when deviations are detected.

The proposed Adaptive RL framework was successfully deployed on a Unitree Go1 robot and evaluated across diverse terrains, including flat ground, slopes, and stairs, under both varying static payloads and dynamic payload scenarios, such as freely moving iron balls placed in a tray mounted on the robot's base. Across all tested conditions, our policy consistently outperformed the baseline controller in terms of accurately tracking both the commanded body height and velocity, demonstrating superior adaptability and robustness to payload variations and environmental changes.

\bibliographystyle{ieeetr}
\footnotesize
\bibliography{references}

\end{document}